%% file: acl_latex.tex
\pdfoutput=1

\documentclass[11pt]{article}

\usepackage[final]{acl}

\usepackage{subcaption}

\usepackage{hyperref}
\usepackage{times}
\usepackage{latexsym}
\usepackage{multirow}
\usepackage{booktabs}
\usepackage{microtype}
\usepackage{multirow}
\usepackage{latexsym}
\usepackage{courier}
\usepackage{amsmath}
\usepackage{booktabs} 
\usepackage{siunitx} 
\usepackage{placeins}
\usepackage{tikz}
\usepackage{hyperref}
\usepackage{float}

\usepackage[T1]{fontenc}

\usepackage[utf8]{inputenc}

\usepackage{microtype}

\usepackage{inconsolata}
\usepackage{soul}

\title{IITK at SemEval-2024 Task 10: Who is the speaker? Improving Emotion Recognition and Flip Reasoning in Conversations via Speaker Embeddings}

\author{Shubham Patel\thanks{\ \ Equal Contributions} \qquad Divyaksh Shukla\footnotemark[1] \qquad Ashutosh Modi \\
Indian Institute of Technology Kanpur (IIT Kanpur)\\
\texttt{\{devang21, divyakshs23\}@iitk.ac.in} \\
\texttt{\{ashutoshm\}@cse.iitk.ac.in} 
}

\begin{document}
\maketitle

\input{sections/abstract}
\input{sections/introduction}
\input{sections/background}
\input{sections/task}
\input{sections/system}
\input{sections/experiments}
\input{sections/results}

\input{sections/conclusion}


\bibliography{custom}

\appendix
\input{sections/appendix}

\end{document}

%% file: sections/abstract.tex
\begin{abstract}

This paper presents our approach for the SemEval-2024 Task 10: Emotion Discovery and Reasoning its Flip in Conversations. For the Emotion Recognition in Conversations (ERC) task, we utilize a masked-memory network along with speaker participation. We propose a transformer-based speaker-centric model for the Emotion Flip Reasoning (EFR) task. We also introduce \textit{Probable Trigger Zone}, a region of the conversation that is more likely to contain the utterances causing the emotion to flip. For sub-task 3, the proposed approach achieves a 5.9 (F1 score) improvement over the task baseline. The ablation study results highlight the significance of various design choices in the proposed method.

\end{abstract}

%% file: sections/introduction.tex
\section{Introduction} \label{sec:intro}




%
%
Conversations between participants carry information that evokes emotions. Emotions include personality, character, temper, and inspiration as the primary psychological parameters that drive them \cite{emotions-review}. Analyzing emotions through language helps uncover the interpersonal sentiments in a conversation at a finer level. This can help build better affective generative models \cite{goswamy-etal-2020-adapting}, like chatbots that understand emotion and respond according to a person's behaviors and personality \cite{meld-fr,colombo-etal-2019-affect}. 



The SemEval-2024 Task 10 \cite{kumar2024semeval} aims at Emotion Recognition (ERC), sub-task 1, and Emotion Flip Reasoning (EFR), sub-tasks 2 and 3, in conversations for two languages, namely English and Hindi-English Code-Mixed. ERC refers to identifying the emotion of different utterances. EFR is about identifying those utterances in the dialogue that caused the emotion of a speaker to change. 

We build upon the models presented in \citet{meld-fr} for ERC and EFR. A speaker's personality is likely to influence the emotions developed in other participants \cite{hazarika-etal-2018-icon}. This inspired us to include information regarding speaker participation to improve the analysis of the emotion of an utterance in conversations. Additionally, for Emotion Flip Reasoning, we propose the Probable Trigger Zone (PTZ), a region of the conversation more likely to consist of the utterance that caused an emotional change in the target participant. This helps us filter out significant non-trigger utterances, reducing the skew in the data. We utilize pre-trained models for computing text embeddings to obtain better representations of utterances.

In sub-task 1, we achieved a weighted F1 score of 45 and 9th rank. For sub-tasks 2 and 3, we secured 5th and 10th position with F1 scores of 56 and 60, respectively. The top scores for each sub-task were 78, 79, and 79, respectively. For sub-task 3, our model improves 5.9 F1 over the baseline model presented in \citet{meld-fr}. The proposed changes have assisted in improving the performance of the system. A limitation of our model is knowing speakers. It might not be possible in all circumstances that this information is available. Also, despite trying to reduce the skew in the data, our model's performance was still impacted. Our models and code can be found here.\footnote{\url{https://github.com/Exploration-Lab/IITK-SemEval-2024-Task-10-Emotion-Flip}}


%% file: sections/background.tex


\section{Related Work} \label{sec:background}
%
%
\subsection{ERC}
The task of emotion prediction has been of active interest in recent years \cite{witon-etal-2018-disney,kumar-etal-2020-baksa,keswani-etal-2020-iitk-semeval,gargi-definitions-acii-2021,gargi-fine-grained-tfacc-2023,singh-etal-2021-end}, including the development of models like ICON \cite{hazarika-etal-2018-icon}, COGMEN \cite{joshi-etal-2022-cogmen}, Instruct-ERC \cite{lei2023instructerc} and the models by \citet{meld-fr}. Also, there has been active research in affective text generation \cite{goswamy-etal-2020-adapting}. Several datasets exist \cite{masac-dataset,poria-etal-2019-meld, Busso2008Dec} that use one or more additional emotions along with Ekman's scheme \cite{paul-ekman-emotion} of emotion representation via six emotion classes, namely, \textit{fear}, \textit{anger}, \textit{joy}, \textit{sad}, \textit{disgust}, and \textit{surprise}.

\citet{hazarika-etal-2018-icon} and \citet{li2020hierarchical} highlight the importance of inter and intra-speaker interactions in a conversation. \citet{li2020hierarchical} achieves this by using three separate transformer-encoder blocks: (1) Conventional masking: masked multi-head self-attention, (2) Intra-speaker masking: all utterances from other speakers are masked, and (3) Inter-speaker masking: all utterances from the current speaker are masked. While this captures relationships, it does not capture the speaker's personality or presence. \citet{hazarika-etal-2018-icon} also considers speakers, but it was modeled on the IEMOCAP dataset \cite{Busso2008Dec} that contains only two participants.

Shapes of Emotion \cite{bansal-etal-2022-shapes}, ICON \cite{hazarika-etal-2018-icon} and its derived model  ERC\_MMN \cite{meld-fr} proposed the concept of speaker-level outputs, which means that during conversational flow, there is a speaker-level GRU to encode the currently spoken utterance. They achieve this by storing vectors representing each speaker and updating them using the speaker-level GRUs' hidden outputs, which are initialized to 0 during the start of a dialogue.

COGMEN \cite{joshi-etal-2022-cogmen} introduces the concept of graphs to conversation flow for emotion recognition. They represent a graph in which each utterance is a node and is related to past or future utterances of the same or different speaker within a time window. CORECT \cite{Nguyen2023corect} leverages on COGMEN and introduces speaker embeddings from MMGCN \cite{wei2019mmgcn} to encode each speaker in the conversation for graph-based interaction and pairwise cross-modal feature interaction.


%
%
\subsection{EFR}
\citet{meld-fr} introduces the relatively new Emotion-Flip Reasoning (EFR) task, which aims to identify past utterances in a conversation that have triggered one’s
emotional state to flip at a certain time. The task of Emotion-Cause Pair Extraction (ECPE) \cite{xia-ding-2019-emotion} and Emotion Cause Extraction (ECE) \cite{gui-etal-2016-event} are similar to EFR, but they aim to extract the causes of emotions from a given text instead of conversations. \citet{meld-fr} present a transformer-based model for EFR and also measure the performance of baseline models CMN \cite{hazarika-etal-2018-conversational}, ICON \cite{hazarika-etal-2018-icon}, DGCN \cite{ghosal-etal-2019-dialoguegcn}, AGHMN \cite{jiao2019realtime}, and Pointer Network \cite{vinyals2017pointer}.



%
%
\subsection{Embeddings}
The performance of models on tasks is influenced by the quality of text representation it uses \cite{Asudani2023Sep}. \citet{nayak-joshi-2022-l3cube} release HingBERT, a BERT model that has been fine-tuned on Hindi-English Code-Mixed corpus. \citet{muennighoff-etal-2023-mteb} introduce the Massive Text Embedding Benchmark (MTEB), which evaluates the performance of text embeddings through different tasks across several datasets. One of the top performers, the \verb|voyage-embeddings|\footnote{\url{https://docs.voyageai.com/embeddings/}}, utilize neural-net models to encode the text into text embeddings.

%% file: sections/task.tex
\section{Task} \label{sec:task}

%
%
%
%
SemEval-2024 Task 10: ``\textbf{E}motion \textbf{Di}scovery and its \textbf{Re}asoning it \textbf{F}lip in Conversations'' \cite{kumar2024semeval}, EDiReF, consisted of three sub-tasks:
\begin{enumerate}
\item ERC in Hindi-English Code-Mixed.
\item EFR in Hindi-English Code-Mixed.
\item EFR in English.
\end{enumerate}

Emotion Recognition in Conversations (ERC) is classifying the utterances in a dialogue into one of the given emotion categories. An emotion flip is said to have occurred when a speaker's utterance differs from his/her previous utterance's emotion. Emotion Flip Reasoning (EFR) refers to identifying the utterances (triggers) that caused an emotional flip. These utterances could have been spoken either by the speaker himself or someone else. For the task of ERC, given the utterances in the conversation and corresponding speaker names, the emotion label for each utterance has to be predicted. For the task of EFR, the emotion labels of utterances have also been provided, and the triggers for a given emotion flip have to be predicted.

%
%
%
%
\begin{table}[t]
    \centering
        \begin{tabular}{  l  r  r  r  }
            \toprule
            \textbf{Sub-task} & \textbf{1} & \textbf{2} & \textbf{3} \\ 
            \midrule
            Emotions & 8 & 8 & 7 \\ 
            Episodes & 343 & 452 & 833 \\ 
            Utterances & 8506 & 11260 & 8747 \\ 
            Triggers & - & 6542 & 5575 \\
            \bottomrule
        \end{tabular}
    \caption{Statistics of the Training Dataset.} 
    \label{tab:basic-stats}
\end{table}

\begin{figure}[t]
    \centering
    \begin{subfigure}{\columnwidth}
        \includegraphics[width=\columnwidth]{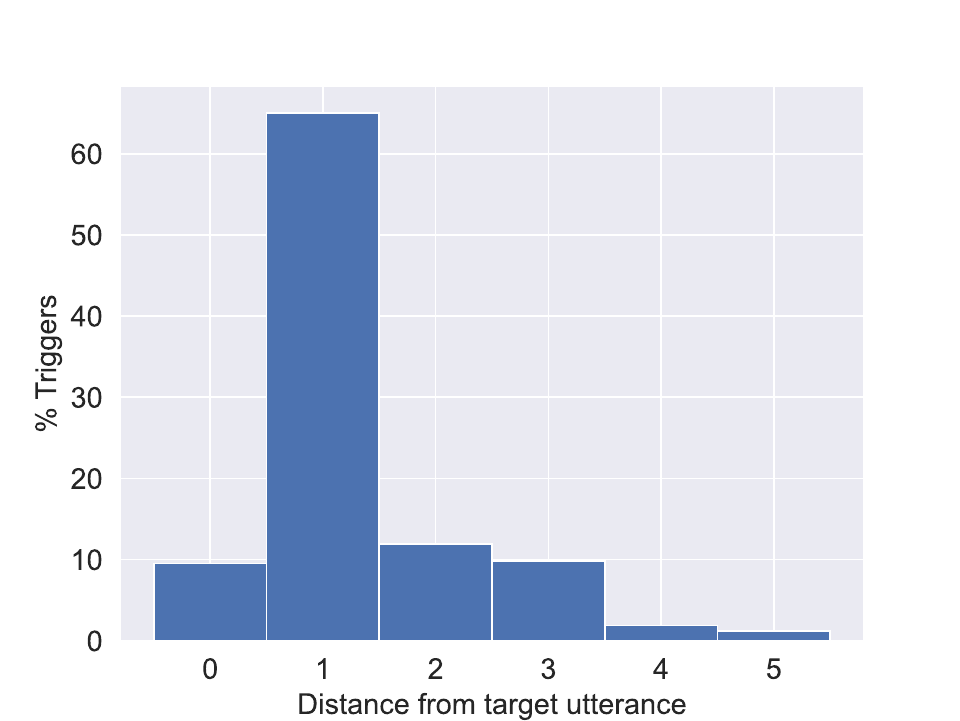}
        \caption{Sub-Task 2.}
        \label{fig:masac-distance}
    \end{subfigure}
    \begin{subfigure}{\columnwidth}
        \includegraphics[width=\columnwidth]{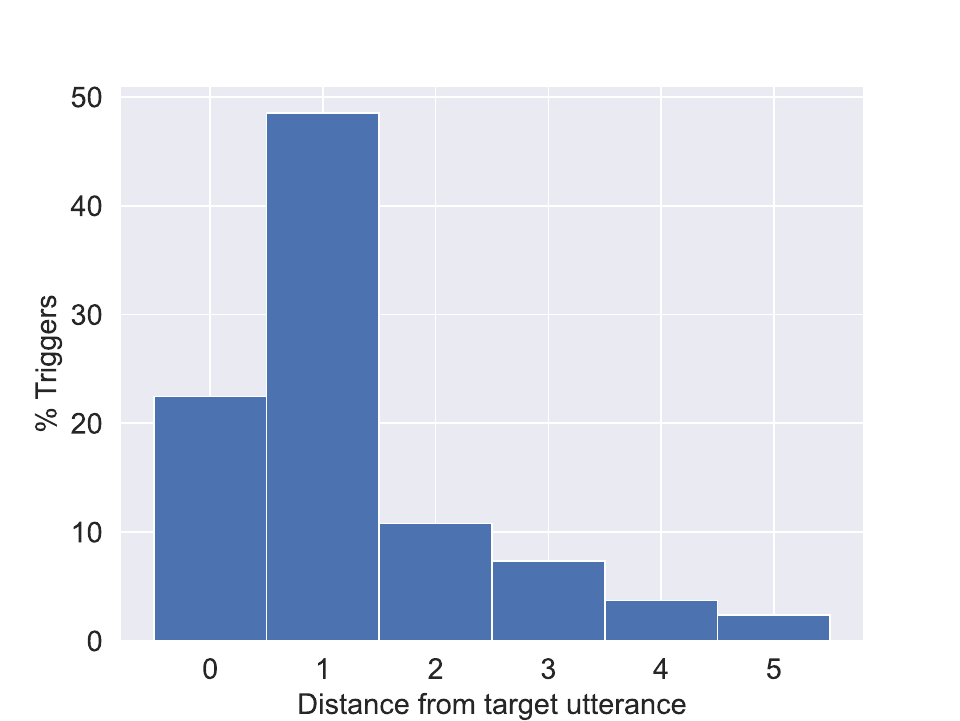}
        \caption{Sub-Task 3.}
        \label{fig:meld-distance}
    \end{subfigure}
    \caption{Distribution of the distance between the target
utterance and the causal utterance for emotion flip.}
\end{figure}

%
%
%
%
 Sub-Tasks 1 and 2 tailor the Hindi-English Code-Mixed dataset - MaSaC \cite{masac-dataset}. The training dataset consists of utterances in Roman script (e.g., ``\textit{yah plastic ke stickers tumne kahan se khariden?}''). Sub-Task 3 uses the MELD-FR dataset presented in \citet{meld-fr} built upon the MELD dataset \cite{poria-etal-2019-meld}. Table \ref{tab:basic-stats} highlights the overall statistics regarding the training set for each sub-task. Figure \ref{fig:masac-distance} and Figure \ref{fig:meld-distance} show the distribution of the triggers as a function of the distance from the target utterance for the sub-task 2 and sub-task 3 training datasets, respectively. 
 

%% file: sections/system.tex
\section{System overview} \label{sec:system}
Inspired by the use of memory networks by \citet{hazarika-etal-2018-icon} and \citet{meld-fr} for emotion recognition, in \ref{sec:erc} we present our model for the task of ERC, sub-task 1. Inspired by a transformer-based \cite{Vaswani-2017} approach for emotion flip reasoning presented by \citet{meld-fr}, in \ref{sec:efr} we present our models for sub-tasks 2 and 3.
    
    





%
%
%
%
\subsection{Utterance Embeddings}
We utilize pre-trained models to compute representations of the utterances in the conversation. Sub-tasks 1 and 2 required the computation of utterance embeddings for code-mixed Hindi-English sentences. We utilized HingBERT to compute the utterance embedding as an average of all the token embeddings in an utterance. Sub-Task 3 consists of utterances in English. We referred to the Massive Text Embedding Benchmark to determine an efficient method to compute utterance embeddings. We experimented with the embeddings presented in \citet{uae-embed} and voyage-embeddings, out of which the latter performed better. Hence, we used the \verb|voyage-lite-02-instruct| model with \verb|query_type| as a \verb|document|.

%
%
%
%
\subsection{ERC}\label{sec:erc}
\begin{figure}[t]
    \centering
    \includegraphics[width=\columnwidth]{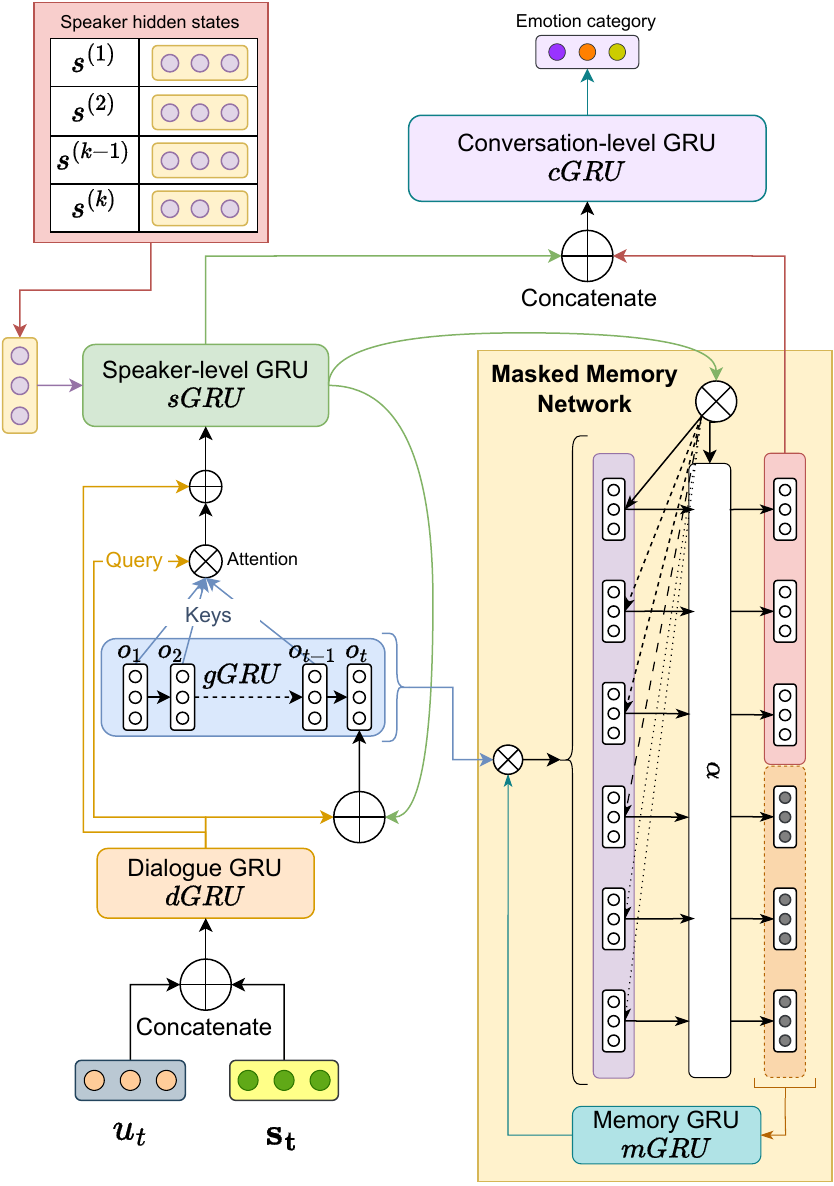}
    \caption{Masked Memory Network with Speaker-Embeddings concatenated with utterance embeddings. Speaker-embeddings are one-hot vectors of 6-dimensions which store 1 at the index of the top-6 speakers, otherwise 0.}
    \label{fig:model-architecture-erc}
\end{figure}

We take inspiration from the Masked Memory Network architecture presented by \citet{meld-fr} and speaker-specific GRUs proposed by \citet{meld-fr} and \citet{hazarika-etal-2018-icon}. We used HingBERT to encode each utterance and then pass them through a dialog-level GRU and a speaker-level GRU. The vectors from the global-level GRUs are passed through a memory network through multiple hops (a cycle of reading from memory and writing back to memory is called a hop. The output is taken from the final memory read operation.) Then, attention is computed between the memory and speaker-level outputs while masking future utterances and concatenating with speaker-level outputs to compute conversation-level outputs. Finally, the obtained features are passed through a trainable linear layer for predicting emotion class. Figure \ref{fig:model-architecture-erc} shows the model architecture. 

\noindent \textbf{Notations:} $u_t$ denotes the embedded utterance at time $t$ in a dialogue, while $s^{(k)}_t$ denotes the $k^{th}$ speaker embedding for utterance $u_t$. $attention(q, k, v)$ is the attention operator applied on query $q$, key $k$ and value $v$. $r$ is the number of hops in the memory layer, inspired from \citet{hazarika-etal-2018-icon}.

\noindent \textbf{Dialogue-level GRU $dGRU$:} This recurrent unit gives a dialogue-level representation of the $u_t$ and gives output as $do_t$. $$do_t = dGRU(u_t \oplus s_{kt})$$

\noindent \textbf{Global-level GRU $gGRU$:} This recurrent unit gives a global-level representation of the utterances $u_{(1:t)}$ till time step $t$, as $o_{(1:t)}$. $$o_t = gGRU(do_t \oplus so_{t-1})$$

\noindent \textbf{Attention Module:} Attention is computed between $do_t$ as query and value and $o_{(1:t-1)}$ as keys to obtain attention-based context for speaker-level GRU layer. $$attention(do_t, o_{(1:t-1)}, do_t)$$

\noindent \textbf{Speaker-level GRU $sGRU$:} This gives a speaker-level recurrent unit that takes inputs $attention$ and speaker hidden state $s^{(k)}$ (taken from a dictionary of size k) and gives outputs $so_t$. The hidden output replaces the dictionary entry for $s^{(k)}$. At the start of a dialogue, the dictionary is empty, and the default hidden state for a new speaker is a zero vector. $$so_t = sGRU(attention(do_t, o_{(1:t-1)}, do_t) + do_t)$$

\noindent \textbf{Masked-Memory Attention:} A memory vector, which represents the previous dialogues and utterances, is obtained by passing $o_{(1:t)}$ through a memory GRU ($mGRU$). This then goes through masked attention with the $so_t$ while masking future utterances and a softmax activation $\alpha$ to give attention weights to each utterance in $o_{(1:t-1)}$. This is then used to update the memory vectors via the $mGRU$ and is concatenated with $so_t$ as an input to $cGRU$. 

\begin{align*}
    temp &= mGRU(o_{(1:t)}) \\
    mem^r &= masked\_attention(temp, so_t) \\
    temp &= mGRU(mem^{r-1}) \\
\end{align*}

\noindent \textbf{Conversation-level GRU $cGRU$:} This layer represents the conversation flow of the dialogue and takes inputs as the concatenation of $so_t$ and masked attention output, to give conversation-level features.$$co_t = cGRU(mem^r+so_t)$$

Finally, the outputs of the $cGRU$ are used to compute the emotion class.$$e_t = W.co_t + b$$




%
%
%
%
\subsection{EFR}\label{sec:efr}

%
%
%
%
\subsubsection{Baseline}
In \citet{meld-fr}, the authors propose a transformer-based model for the task of EFR, whose architecture is as follows. Utterance embeddings for each utterance are computed using BERT \cite{devlin-etal-2019-bert}. The utterance embeddings of a conversation are passed through a transformer to compute a contextualized utterance embedding for each utterance. The emotion classes for each utterance are encoded as a one-hot vector and passed through a GRU to compute the emotion-history vector. For each utterance, its contextualized embedding, the contextualized embedding of the target utterance, and the emotion-history vector are passed through a linear layer to make a prediction.

%
%
%
%
\subsubsection{Speaker-Aware Embeddings}
As highlighted by \citet{li2020hierarchical} and \citet{hazarika-etal-2018-icon}, speaker interaction also drives the emotion of an utterance. Unlike their approach for modeling intra and inter-speaker interaction, we believe that the participation of certain speakers in the conversation drives the flip in the emotion of an utterance. Providing information regarding the speaker will help the model learn the nature of the specific speakers. To incorporate this, we utilize that the speakers in the test and the train set overlap. 

An aspect of the conversation that has not been captured by the baseline model regarding the speakers of an utterance. In the baseline model, each utterance is treated as an independent text, and its embedding has been computed. This has failed to incorporate the information regarding who the speaker of a given utterance was. To incorporate this aspect, we concatenate the utterance embeddings with a one-hot vector denoting the speaker to create speaker-aware embeddings. This equips the model with the ability to capture the behavioral trends of specific speakers. 

%
%
%
%
\subsubsection{Probable Trigger Zone (PTZ)}
We propose a hypothesis regarding the possible location of triggers. We divide the conversation into two parts. The first part consists of the utterances before the target speaker's previous utterance. The second part consists of utterances from the target speaker's previous utterance to his target utterance. We call the second part the Probable Trigger Zone (PTZ). 

We hypothesize that no triggers lie in the first part of the conversation. Since the target speaker's emotion has flipped during the second part of the conversation, it is more likely that the causes for the emotion flip lie in the second part. Suppose the trigger causing the target emotion had been in the first part. In that case, it is more likely that it would have already caused the emotion of the previous utterance of the target speaker to flip. Then, the same emotion would have been carried to the target utterance, wrongly implying that no emotion flip occurred at the target utterance. To incorporate the hypothesis, we mask any predictions made by the model outside the Probable Trigger Zone. In section \ref{sec:experiments}, we discuss how PTZ  helps to reduce skew in the dataset.

\begin{figure}[t]
    \centering
    \includegraphics[width=\columnwidth]{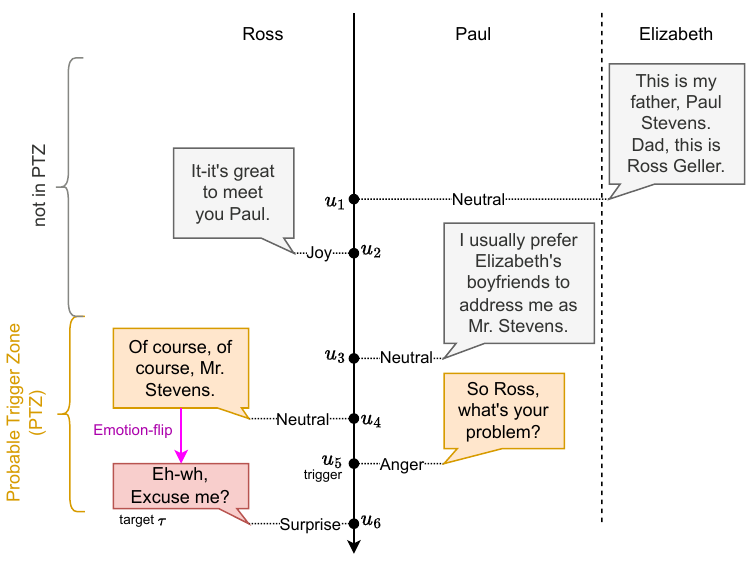}
    \caption{Probable Trigger Zone.}
    \label{fig:ptz}
\end{figure}

For example, consider the conversation in Figure \ref{fig:ptz}. Here, the target speaker is \textit{Ross} with the target utterance $u_6$ and his previous utterance $u_4$. The probable trigger zone consists of utterances from $u_4$ to $u_6$. Due to the ``surprise-causing'' statement $u_5$ in PTZ, \textit{Ross's} emotion flips from \textit{Neutral} to \textit{Surprise}. If this ``surprise-causing'' statement had been present outside the PTZ, i.e., before the previous utterance $u_4$, then the emotion of $u_4$ would likely have been \textit{Surprise}.

%
%
%
%
\subsubsection{Emotion-Aware Embeddings}
Using an Emotion-GRU, the baseline model computes an emotion-history vector from the emotion labels. It uses this emotion-history vector in the final linear layer to predict the utterance label. A possible shortcoming of the above is that the linear layer has access only to the emotion history rather than to the emotion labels of the individual utterances. Also, the transformer layer cannot access the emotion labels while computing the contextualized utterance embeddings. Providing those to the transformer will also allow the embeddings to be emotion-aware. We concatenate our speaker-aware utterance embeddings and one-hot emotion labels to incorporate the above.

%
%
%
%
\subsubsection{Model Functioning}
\begin{figure}[t]
    \centering
    \includegraphics[width=\columnwidth]{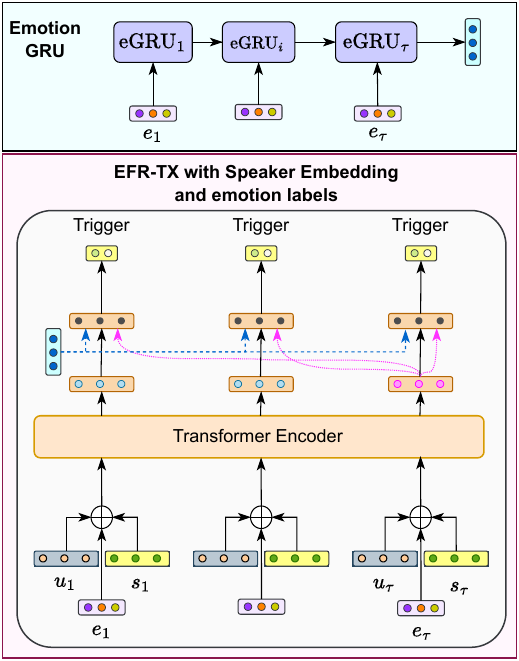}
    \caption{Architecture of the model proposed for the task of Emotion Flip Recognition.}
    \label{fig:model-architecture-efr}
\end{figure}

Figure \ref{fig:model-architecture-efr} presents the model architecture used for the task of EFR. The target utterance is denoted by the subscript $\tau$. Each utterance $u_{t}$ of a dialogue $d$ is concatenated with its true emotion label $e_t$ and one-hot speaker embedding $s_{t}$. This is then passed through the transformer to take into account the context. The Emotion-GRU computes the emotion-history vector. For each utterance, its and the target utterance's contextualized representation and the emotion-history vector are passed through a linear layer to make the prediction.

%% file: sections/experiments.tex
\section{Experimental setup} \label{sec:experiments}





%
%
%
%
\subsection{Training Details}
For sub-task 1, we chose a sequence length $seq\_len$ of $15$, i.e., we break a long conversation into disjoint smaller conversations with utterances less than equal to $seq\_len$. For sub-tasks 2 and 3, we use a window size $w$ of $5$. We consider only the last $w$ utterances in a conversation to predict the trigger, i.e., $U_{n-w+1}, U_{n-w+2}... U_{n-1}, U_{n}$. Table \ref{tab:hyperparams} contains details of the hyperparameters we used to train the models. To limit the size of the vector denoting the speaker, keep retained information regarding the top $k=6$ speakers. We chose the top $6$ speakers since this covered nearly $80-85\%$ of utterances in the corpus.
\begin{table}[t]
    \centering
    \begin{tabular}{lccc}
         \toprule
         \textbf{Sub-Task} & \textbf{1} & \textbf{2} & \textbf{3} \\
         \midrule
         Embedding Size & 768 & 768 & 1024 \\
         Batch Size & 64 & 2000 & 1000 \\
         Learning Rate & 1e-04 & 5e-07 & 5e-07 \\
         Weights & Inv Sqrt & Inv & Inv \\
         Epochs & 100 & 1000 & 1000 \\
         Best Epoch & 80 & 299 & 549 \\
         Training Time & 10 hr & 3 hr & 3 hr \\
         \bottomrule
    \end{tabular}%
    \caption{Hyperparameters for each of the sub-tasks. \small{Weights refers to the weights in the cross entropy loss. \textit{Inv}: Inverse of the supports. \textit{Inv Sqrt}: Inverse of the square root of the supports.}}
    \label{tab:hyperparams}
\end{table}
We used Adam optimizer \cite{kingma2017adam} for all the sub-tasks, with a weight decay of 1e-5. Training of models has been done using Kaggle\footnote{\url{https://www.kaggle.com/}} P100 GPUs.

%
%
%
%
\subsection{Effect of Hypothesis and Sequence Length}\label{subsec:5.3}
In Table \ref{tab:hypo_stat_2} and Table \ref{tab:hypo_stat_3}, we highlight the impact of the hypothesis and selection of sequence length on the datasets. On reducing the window size $w$ to $5$, a significant number of negative labels have been eliminated, while there has not been much impact on the number of positive labels. Applying the hypothesis has helped mitigate the skew in the data, although there has been a slight impact on the number of positive labels. \textit{Setting 1} refers to considering only the utterances within the window size $w=5$. \textit{Setting 2} refers to considering utterances that are both within the window and in the probable trigger zone.
\begin{table}[t]
    \centering
    \begin{tabular}{lccc}
        \toprule
         \textbf{Dataset} & \textbf{0} & \textbf{1} & \textbf{Ratio} \\
        \midrule
        Original & 92233 & 6544 & 14.1 \\
        Setting 1 & 17539 & 6425 & 2.7 \\
        Setting 2 & 11535 & 5839 & 2.0 \\
        \bottomrule
    \end{tabular}%
    \caption{Effect of PTZ on Dataset, Sub-Task 2. \\
    \small{\textit{Setting 1} and \textit{Setting 2} as defined in Section \ref{subsec:5.3}.}}
    \label{tab:hypo_stat_2}
\end{table}

\begin{table}[h]
    \centering
    \begin{tabular}{lccc}
    \toprule
        \textbf{Dataset} & \textbf{0} & \textbf{1} & \textbf{Ratio} \\
        \midrule
        Original & 29416 & 5575 & 5.3 \\
        Setting 1 & 13483 & 5177 & 2.6 \\
        Setting 2 & 7834 & 4542 & 1.7 \\
        \bottomrule
    \end{tabular}%
    \caption{Effect of PTZ on Dataset, Sub-Task 3. \\
    \small{\textit{Setting 1} and \textit{Setting 2} as defined in Section \ref{subsec:5.3}.}}
    \label{tab:hypo_stat_3}
\end{table}

%% file: sections/results.tex
\section{Results} \label{sec:results}

For Sub-Task 1, the dataset consisted of a non-uniform distribution of labels, with \textit{neutral} being the most frequent. A model that predicts the emotion category of each utterance to be \textit{neutral} achieves a weighted F1 of $24.36$. We consider this as a simple \textit{neutral} baseline for sub-task 1. For Sub-Task 2, we have kept the baseline as a rule-based model that predicts the previous utterance to be a trigger and the rest of all utterances non-triggers. The data for the second sub-task is highly skewed as can be seen in Figure \ref{fig:masac-distance}. Due to this baseline performs exceptionally well, as can be observed in Table \ref{tab:baselines}. For Sub-Task 3, we use ERC\textsuperscript{True} EFR-TX from \citet{meld-fr} as the baseline.

\begin{table}[h]
    \centering
    \begin{tabular}{cccc}
         \toprule
         \textbf{Sub-Task} & \textbf{Model} & \textbf{Metric} & \textbf{Value} \\
         \midrule
         2 & Rule-Based & F1 & 79.15 \\ 
         3 & $\mathrm{ERC}^{\mathrm{True}}$ EFR-TX & F1 & 53.9 \\ 
         \bottomrule
    \end{tabular}%
    \caption{Baselines for various Sub-Tasks.\\
    \small{Rule-Based: A rule-based model that predicts the previous utterance as a trigger and the rest as non-triggers.}}
    \label{tab:baselines}
\end{table}
%
%
%
%
\subsection{Model Performance}
We have highlighted the performance of our models on the test data in Table \ref{tab:results}. For sub-task 2, we get precision and recall scores of 0.73 and 0.83, respectively. For sub-task 3, we get precision and recall scores of 0.74 and 0.80, respectively.

\begin{table}[t]
    \centering
    \resizebox{\columnwidth}{!}{
    \begin{tabular}{ccccc}
         \toprule
         \textbf{Sub-Task} & \textbf{Metric} & \textbf{Our} & \textbf{Best} & \textbf{Rank} \\
         \midrule
         1 & Weighted F1 & 44.80 & 78 & 9 \\  
         2 & F1 & 56.35 & 79 & 5 \\  
         3 & F1 & 59.78 & 79 & 10 \\  
         \bottomrule
    \end{tabular}%
    }
    \caption{Results on the Test Set.}
    \label{tab:results}
\end{table}
%
%
%
%

%
%
\subsection{Error Analysis}
\begin{figure}[h]
    \centering
    \includegraphics[width=\columnwidth]{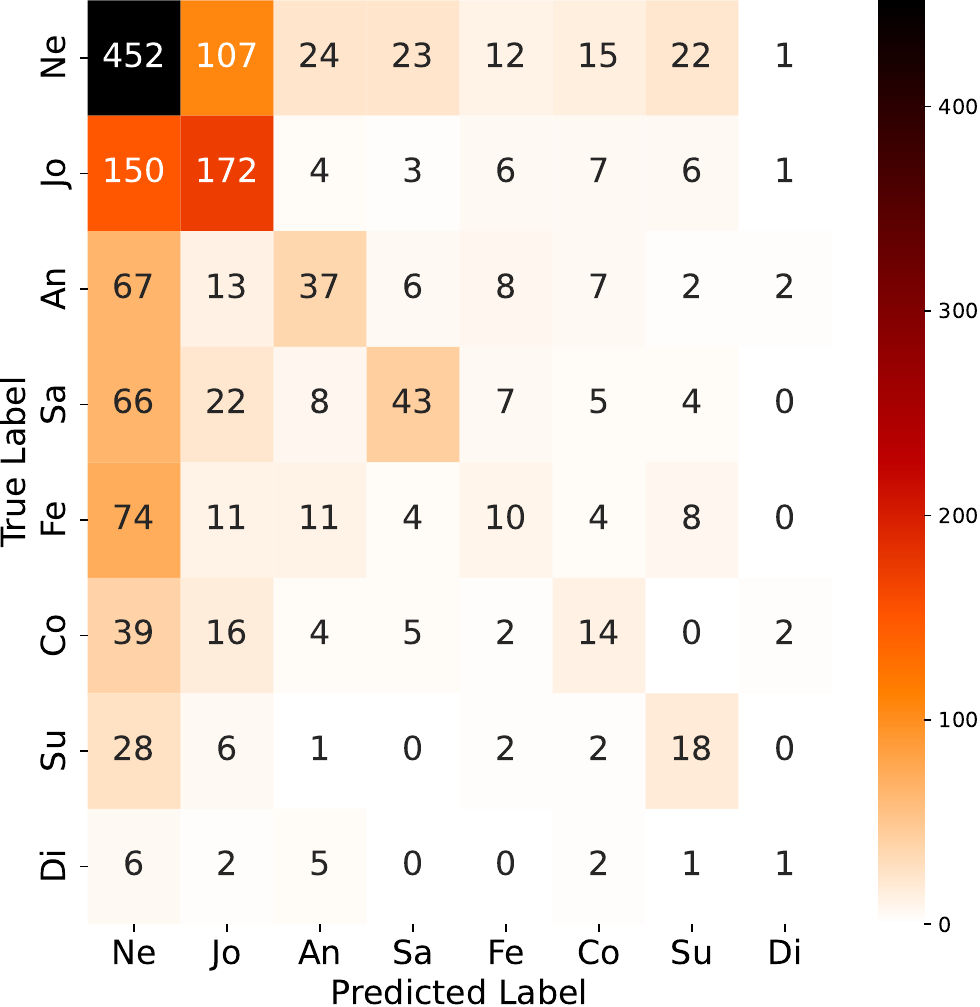}
    \caption{Confusion Matrix for Sub-Task 1.} 
    \label{fig:con_mat_1}
\end{figure}

For sub-task 1 and sub-task 3, our model performed better than the baselines, but not for sub-task 2. For sub-task 1, the dataset consisted of a non-uniform distribution of labels in the training dataset. Due to this skew in the data, the model has shown different performances for different labels. The predictions for sub-task 1 have been highlighted in Figure \ref{fig:con_mat_1}. For EFR, the usage of a window size $w=5$ utterances has helped to eliminate a large number of non-triggers. Due to this, the model's predictions have many true negatives, as can be seen in Table \ref{tab:con_mat_2} and Table \ref{tab:con_mat_3}. But despite this, there was still skew in the data, which impacted the model's performance in predicting the minority class of triggers. The data for sub-task 2 is highly skewed, as can be seen in Figure \ref{fig:masac-distance}. We suspect this is why our model has performed poorer than the baseline. 

\begin{table}[h]
    \centering
    \resizebox{0.5\columnwidth}{!}{
    \begin{tabular}{cc|rr}
    \toprule
        \multicolumn{2}{c}{} & \multicolumn{2}{c}{\textbf{Predicted}} \\ 
        \multicolumn{2}{c}{} & \textbf{0} & \textbf{1} \\ 
        \hline
        \multirow{2}{*}{\textbf{True}} & \textbf{0} & 6943 & 331 \\ 
         & \textbf{1} & 123 & 293 \\
    \end{tabular}%
    }
    \caption{Confusion Matrix Sub-Task 2.}
    \label{tab:con_mat_2}
\end{table}

\begin{table}[h]
    \centering
    \resizebox{0.5\columnwidth}{!}{
    \begin{tabular}{cc|rr}
    \toprule
        \multicolumn{2}{c}{} & \multicolumn{2}{c}{\textbf{Predicted}} \\ 
        \multicolumn{2}{c}{} & \textbf{0} & \textbf{1} \\ 
        \hline
        \multirow{2}{*}{\textbf{True}} & \textbf{0} & 6735 & 738 \\ 
         & \textbf{1} & 356 & 813 \\
    \end{tabular}%
    }
    \caption{Confusion Matrix Sub-Task 3.}
    \label{tab:con_mat_3}
\end{table}

%
%
\begin{figure}[t]
    \centering
    \includegraphics[width=\columnwidth]{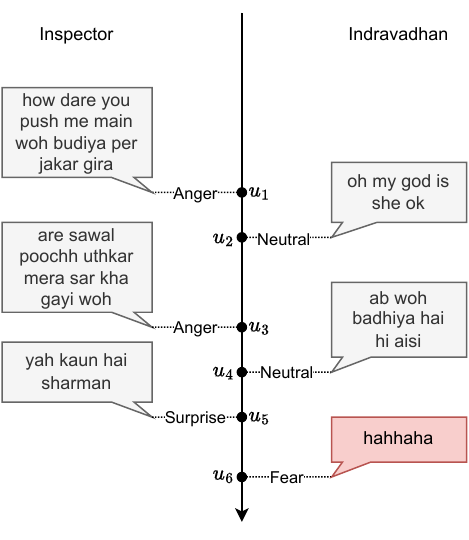}
    \caption{Example of an erroneous emotion labeling from the model. The true label is `Fear,' but the model predicted `Neutral.'}
    \label{fig:error-1}
\end{figure}
Figure \ref{fig:error-1} is an example of an erroneous emotion labeling of the model on the test set of sub-task 1 (ERC). Here, the utterance marked in red has the true label as `Fear,' but the model predicts `Neutral.' This is due to the sharp change in conversation context at $u_5$. Also, `hahhaha' directly corresponds to laughing, but in this case, the speaker at $u_6$ utters `hahhaha' as he is worried that the inspector is looking for `Sharman'. The speaker, Indravardhan, who continuously interacted with the inspector, showed neutral emotion. This led to storing vectors corresponding to neutral for Indravardhan in the memory network, leading to misclassification of emotion at $u_6$.
%
%
\subsection{Ablation Study}
The application of the hypothesis has assisted in removing a few of the wrongly guessed triggers. This has improved the model's performance, as seen in Table \ref{tab:hypo-effect}. We also experimented with another approach of making predictions only inside the PTZ instead of masking the outside labels. This was done by training in the model and making predictions only using the utterances within the probable target zone. In this case, the model's performance was poorer. We suspect this is because the context the model gets in the latter is more restricted than in the first case. Due to this, the model is not able to make predictions effectively. 

\begin{table}[t]
    \centering
    \resizebox{\columnwidth}{!}{
    \begin{tabular}{ccccc}
         \toprule
         \textbf{Sub-Task} & \textbf{Masks} & \textbf{F1 Before} & \textbf{F1 After} & \textbf{Change} \\
         \midrule
         2 & 1 & 56.29 & 56.35 & +0.06 \\
         3 & 78 & 58.68 & 59.78 & +1.10 \\ 
         \bottomrule
    \end{tabular}%
    }
    \caption{Improvements by PTZ.\\
    \textit{Masks}: The number of trigger predictions made by the model outside the Probable Trigger Zone, which had been masked to 0.}
    \label{tab:hypo-effect}
\end{table}

%% file: sections/conclusion.tex
\section{Conclusion} \label{sec:conclusion}


In this paper, we discussed approaches to improve the masked memory network architecture for emotion recognition (ERC) and transformer-based architecture for emotion flip reasoning (EFR) by incorporating speaker information into the embeddings and making better use of the emotion labels. We also employed an approach of focusing the model's prediction on more likely regions to identify triggers by defining the Probable Trigger Zone in conversations. This, along with considering a window of last-few utterances, assisted in reducing the bias in the data. 

\noindent\textbf{Limitations:} Our model assumes that the training and testing data consist of the same speakers. While this would be true for many benchmark datasets of emotion analysis in conversations, it might not be true in all real-world circumstances. Another limitation of the proposed approach is the training time.

\noindent\textbf{Future Work:} In the future, we can include speaker information across dialogues for ERC to capture better speaker semantics by using learnable embeddings for each speaker updated by the hidden outputs of the speaker-level GRU. However, to apply the above, we need to know the number of speakers in the datasets, training, and testing. Additionally, the model becomes dependent on the number of speakers.

A possible approach to help mitigate the assumption of having common speakers and knowing the number of speakers in the training and test time could be exploring further modeling inter and intra-speaker dependencies, as shown in \citet{li2020hierarchical} and \citet{hazarika-etal-2018-icon}. They propose models that capture speaker relationships but are not dependent on the number of speakers.

Mitigating the issues of skewed data can be further explored to enhance the system's performance. Also, addressing other aspects of conversations, such as whom the statement is being told to and treating names of speakers in utterances differently from simple pronouns, can be explored.



%% file: sections/appendix.tex



%